\documentclass{article}
\usepackage[utf8]{inputenc}
\usepackage{amsmath, amssymb}
\usepackage{lipsum}
\usepackage{graphicx}
\usepackage{hyperref}
\usepackage[numbers]{natbib}
\usepackage{geometry}
\usepackage{tikz}
\usepackage{subcaption}
\begin{document}
\title{Control Plane as a Tool: A Scalable Design Pattern for Agentic AI Systems}

\author{
  Sivasathivel Kandasamy \\
  \texttt{sivasathivel@yahoo.com} \\
}

\maketitle

\begin{abstract}
Agentic AI systems represent a new frontier in artificial intelligence, where agents—often based on large language models (LLMs)—interact with tools, environments, and other agents to accomplish tasks with a degree of autonomy. These systems show promise across a range of domains, but their architectural underpinnings remain immature. This paper conducts a comprehensive review of the types of agents, their modes of interaction with the environment, and the infrastructural and architectural challenges that emerge. We identify a gap in how these systems manage tool orchestration at scale and propose a reusable design abstraction: the “Control Plane as a Tool” pattern. This pattern allows developers to expose a single tool interface to an agent while encapsulating modular tool routing logic behind it. We position this pattern within the broader context of agent design and argue that it addresses several key challenges in scaling, safety, and extensibility.
\end{abstract}

\section{Introduction}

Agents in software are not a new concept. The foundational definition can be traced back to Wooldridge and Jennings~\cite{wooldridge1995intelligent}, who defined software agents as autonomous, goal-directed computational entities capable of perceiving and acting upon their environment. Historically, such agents have been explored across domains like robotics, multi-agent systems, and distributed computing.

The advent of generative AI—especially large language models (LLMs) such as GPT-4~\cite{openai2023gpt4}, Claude~\cite{anthropic2023claude}, and Gemini~\cite{google2024gemini}—has dramatically transformed this paradigm. LLM-driven agents are no longer bound by pre-coded rules; they now exhibit emergent reasoning, multi-step planning, memory awareness, and flexible tool use. This evolution has given rise to a new class of intelligent systems: \textbf{Agentic AI}.

We define \textbf{Agentic AI} as autonomous software programs, often LLM-powered, that can perceive their environment, plan behaviors, invoke external tools or APIs, and interact with both digital environments and other agents to fulfill predefined goals. These systems are characterized by goal-seeking autonomy, tool adaptability, contextual memory, and multi-agent coordination~\cite{ibm2024agentic, aisera2024agentic, survey2024agentic}.

Agentic AI has rapidly entered mainstream discourse, with organizations seeking to embed agent-based workflows into domains such as customer service, software engineering, and operations. While some cases demonstrate meaningful gains~\cite{bubeck2023sparks}, others are driven by hype cycles and premature generalization~\cite{bommasani2022opportunities}.

The core production value of Agentic AI lies in:
\begin{itemize}
    \item \textbf{Autonomous Decision-Making:} Dynamic task planning and real-time behavioral adaptation.
    \item \textbf{Multi-Tool Integration:} Composition across APIs, search interfaces, and databases.
    \item \textbf{Contextual Reasoning:} Use of memory and history for iterative improvement.
    \item \textbf{Composable Workflows:} Encapsulation of agents as modular, role-oriented microservices.
\end{itemize}

To realize these capabilities, developers rely on a combination of agentic design patterns, including:
\begin{itemize}
    \item \textbf{Reflection Pattern (ReAct)}~\cite{yao2023react}: Alternates between reasoning and acting.
    \item \textbf{Tool Use Pattern}~\cite{deeplearning2024tooluse}: A \textit{tool} can be defined as a piece of code that the Agent uses to observe or act towards achieving its goal. The pattern focuses on agents that uses tools to achieve their goal
    \item \textbf{Hierarchical Agentic Pattern}~\cite{xu2023hierarchical}: Decomposes planning across layered sub-agents.
    \item \textbf{Collaborative Agentic Pattern}~\cite{wu2023autogen, crewai2024}: Assigns roles to specialized agents that cooperate toward a shared objective.
\end{itemize}
In Agentic-AI systems, a \textit{tool} can be defined as a piece of code that the Agent uses to observe or effect change to achieve its goal.
Most production-grade systems employ hybrid designs, mixing multiple patterns to meet business constraints. In parallel, several frameworks have emerged to reduce orchestration complexity and abstract common operations:

\begin{itemize}
    \item \textbf{LangChain}~\cite{langchain2023framework}: A Python framework that chains prompts, tools, and memory components.  
    \textit{Focus:} Prompt-based orchestration and memory integration.  
    \textit{Limitation:} Tight coupling of agent logic and tool invocation leads to brittle workflows.

    \item \textbf{LangGraph}~\cite{langgraph2024}: A graph-based orchestration runtime supporting condition-based tool chains and node-based state handling.  
    \textit{Focus:} Declarative, recoverable workflows.  
    \textit{Limitation:} Requires explicit node wiring and is less dynamic for runtime tool adaptation.

    \item \textbf{AutoGen}~\cite{wu2023autogen}: An LLM-based multi-agent library emphasizing role separation and dialog coordination.  
    \textit{Focus:} Agent-to-agent conversation and memory persistence.  
    \textit{Limitation:} Orchestration is hardcoded; lacks modular tool routing logic.

    \item \textbf{CrewAI}~\cite{crewai2024}: A lightweight framework for role-based multi-agent collaboration.  
    \textit{Focus:} Domain-specialized agents working in crews.  
    \textit{Limitation:} Static role definitions; limited support for dynamic role/tool mutation.

    \item \textbf{Anthropic MCP}~\cite{anthropic2023mcp}: A schema-centric protocol enabling LLMs to securely invoke external tools.  
    \textit{Focus:} Interoperability and tool safety via typed interfaces.  
    \textit{Limitation:} Steep learning curve; orchestration logic is implicit and non-modular.
\end{itemize}

Despite these advancements, productionizing Agentic AI remains challenging due to:
\begin{itemize}
    \item \textbf{Tool Orchestration Complexity:} Scaling APIs without prompt bloat or entangled logic~\cite{qin2023toolllm, deeplearning2024tooluse}.
    \item \textbf{Governance and Observability:} Ensuring traceability and enforcement of tool usage policies~\cite{wu2023autogen, langgraph2024, anthropic2023mcp}.
    \item \textbf{Memory Synchronization:} Maintaining consistent state across workflows~\cite{crewai2024, langchain2023framework}.
    \item \textbf{Cross-Agent Coordination:} Preventing task collisions and misaligned objectives~\cite{wu2023autogen, yao2023react}.
    \item \textbf{Adaptability vs. Safety:} Controlling exploratory behavior while preserving reliability~\cite{bommasani2022opportunities, bubeck2023sparks}.
\end{itemize}

These challenges reveal a deeper architectural design gap. This paper focuses specifically on the challenges related to tool usage by agents. The major contribution of this article is to provide an architectural design pattern that addresses the following limitations in tool handling in Agentic AI systems:
\begin{enumerate}
    \item Add, remove, or modify tools without changing agent code or prompts.
    \item Learn and personalize tool usage for specific tasks and users.
    \item Track tool usage and enforce organizational or compliance policies.
    \item Select tools dynamically based on context or metadata.
    \item Reduce the learning curve for developers building agentic systems.
    \item Enable and simplify distributed, collaborative development of tools across teams.
\end{enumerate}

The next section introduces the proposed architectural pattern—\textit{Control Plane as a Tool}—and discusses how it addresses the identified gaps in tool orchestration within Agentic AI systems. In the following sections, we demonstrate the application of this pattern by designing a simplified chatbot system and outline future directions for extending this work across other facets of Agentic AI architecture.

\section{Proposed Design Pattern: Control Plane as a Tool}
This section introduces a reusable design pattern - \textit{Control Plane as a Tool} - that modularizes and enhances tool orchestration in Agentic AI Systems. The pattern aims to decouple tool management from the Agent's reasoning and decision layers. Thereby enabling flexibility, observability, and scalability across systems. 
Naturally, this pattern can be considered as an extension of the \textit{Tool-use Pattern}. 

\subsection{Design Goals}
The \textit{Control Plane as a Tool} pattern is driven by the following goals:
\begin{itemize}
    \item \textbf{Modularity}: The tool logic should be abstracted from the agent, allowing tools to be modified, added, or removed without changing the agent’s prompt or control logic.
    \item \textbf{Dynamic Selection}: Tool invocation should be dynamic, based on task requirements, metadata, user profiles, or past interactions.
    \item \textbf{Governance and Observability}: Tool usage should be auditable, allowing the enforcement of organizational or safety policies.
    \item \textbf{Cross-Framework Portability}: As a design pattern, it would be framework-agnostic and can be easily used with any framework.
    \item \textbf{Developer Usability}: Developers should be able to use a single tool interface and offload orchestration complexity to the control plane. 
    \item \textbf{Support for Personalization}: Agents should be able to learn and adapt tool selection policies based on feedback or task success.
\end{itemize}

\subsection{Pattern Structure}
In simple Terms, \textit{Control Plane}, in this context, is a piece of software that configures and routes the data between the configured tools and the agents. The set of tools configured forms with Tools Layer and one or more agents configured to the control plane to use the tools form the Agentic Layer.
The Control Plane is exposed to the agent as a \texttt{tool()}, similar to other callable tools (e.g., search, calculator, database). Internally, the control plane executes the following sequence:

\begin{enumerate}
    \item The agent queries the control plane with an intent or query.
    \item Parses metadata of the tools and retrieves relevant candidate tools.
    \item Applies routing logic (e.g. semantic similarity, user context, policy filters, user preference, etc.).
    \item Calls the appropriate tool, and logs the interaction.
    \item Returns the output of the tool to the agent.
\end{enumerate}

\begin{figure}
    \centering
    \begin{subfigure}{.5\textwidth}
        \centering
        \includegraphics[width=0.5\linewidth]{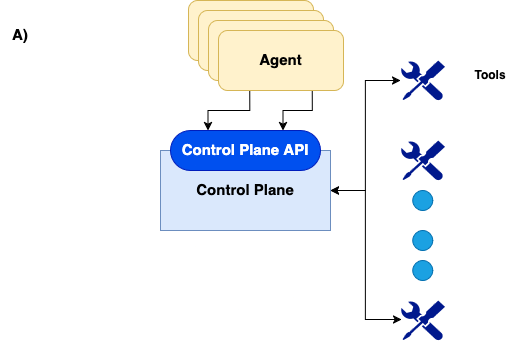}
        \caption{Agents-Tool Separation Through Control Plane}
    \end{subfigure}%
    \begin{subfigure}{.5\textwidth}
        \centering
        \includegraphics[width=0.5\linewidth]{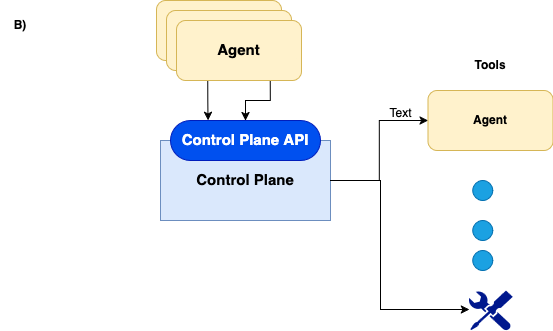}
        \caption{Agents as Tool Through Control Plane}
    \end{subfigure}%
    \caption{Figures show how control plane help with the interaction of agents and tools}
    \label{fig:fig1}
  \end{figure}

This makes orchestration transparent from the agent’s point of view, supporting reuse, caching, validation, and dynamic composition.
Figure \ref{fig:fig1} shows an overview of the high-level Control Plane. Figure \ref{fig:fig1}B also shows that the proposed pattern enables interaction between agents as well through the control plane.
\noindent%

\begin{figure}
    \centering
    \includegraphics[width=0.78\linewidth]{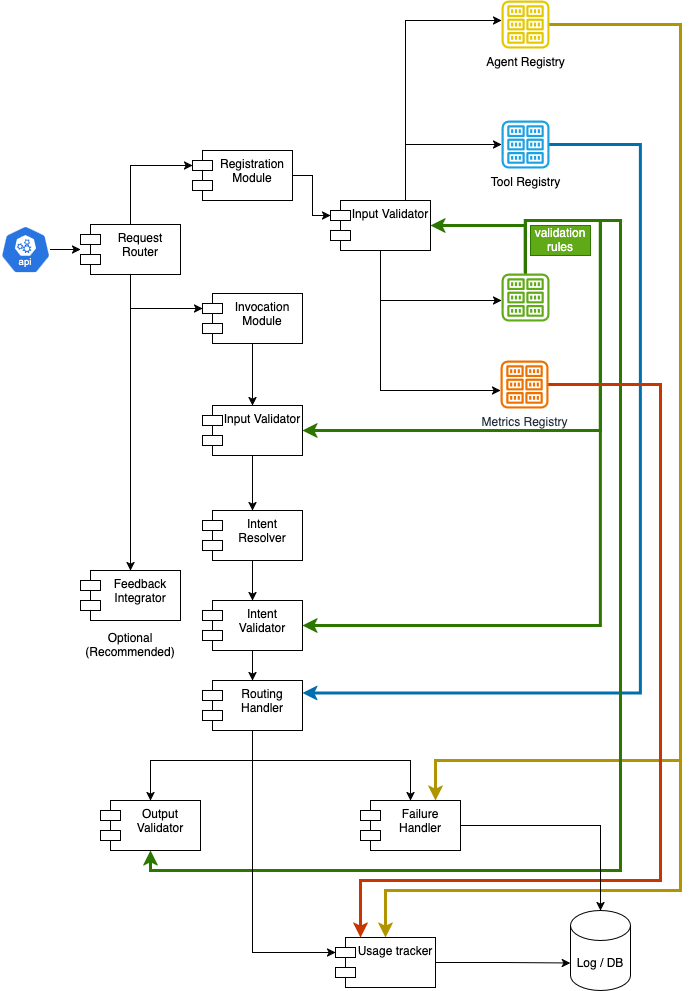}
    \caption{Control Plane Architecture}
    \label{fig: control plane internals}
\end{figure}

The internals of the Control Plane is provided in the Figure \ref{fig: control plane internals}

Agents and externals systems are expected to interact with the control plane through an API endpoint or a CLI. \texttt{Request Router} module decodes the incoming request and route it to the appropriate modules viz \texttt{Registration Module}, \texttt{Invocation Module} and \texttt{Feedback Integration Module}. The main goal of the \texttt{Registration Module} is to register the interacting agents, tools, validation rules and metrics. 

The \texttt{Invocation Module}, module helps the \textit{invoking} agents to query a tool or other registered agents. \texttt{Input Validator} assess the inputs passed for data integrity, safety and alignment, based on the validation rules in the \texttt{validation registry}. Once the inputs are validated, the \texttt{Intent Resolver Module} try to understand the invoking agents' intentions to identify the correct tools/toolset. \texttt{Routing Handler}, based on the resolved intent identify the tools(incl. agents) and their sequence of invocations. Once the outputs from the tools are consolidated, \texttt{output validator }, validates the outputs again to make sure the output complies with registered rules and regulations. Once the results are validated it is returned to the invoking agent. 

The \texttt{Feedback Module}, helps to integrate user feedbacks into the systems so that the tool selection and their sequences can be personalized as per user preference. Though it is an optional module, it highly recommended for performance and accuracy. 

The control plane will be registered as a tool in an agentic,so that each agent has to bind to only one tool, simplifying the process. 

The proposed architecture is considered a design pattern as either be implemented through an Agentic approach or as a set of microservices. Both have the advantages and disadvantages. The non-Agentic approache keeps the complexity to minimum and might be less expensive. However, the Agentic Approach would provide more flexibility and extendability.

\section{Comparison with Model Context Protocol}
The Model Context Protocol (MCP) \cite{anthropic2023mcp} has a similar objective to that of the proposed approach. Hence it become necessary the similarities and differences between the two. The table \ref{tab:cp_mcp_similarities} shows the similarities between the two systems. 
(\paragraph{Disclaimer.} Model Context Protocol (MCP) is a tool interface specification introduced by Anthropic. This paper does not implement, replicate, or reverse-engineer MCP. All comparisons are based on publicly available documentation and are intended solely for academic discussion and architectural contrast.)

\subsection{Similarities Between the Control Plane and MCP}

The proposed Control Plane architecture shares several goals and structural traits with Anthropic's Model Context Protocol (MCP), particularly in its emphasis on safety, tool registration, and structured invocation.

\begin{table}[ht]
\centering
\caption{Similarities Between Control Plane and MCP}
\label{tab:cp_mcp_similarities}
\begin{tabular}{|p{4.5cm}|p{9cm}|}
\hline
\textbf{Feature} & \textbf{Description} \\
\hline
\textbf{Tool Registration} & Both systems require structured metadata or schema registration for external tools. MCP uses JSON schema; the Control Plane maintains a Tool Registry. \\
\hline
\textbf{Input Validation} & Both validate tool inputs using schema constraints. MCP enforces JSON Schema at runtime, while the Control Plane uses a dedicated \textit{Input Validator} module. \\
\hline
\textbf{Invocation Routing} & MCP tools are invoked based on prompt-matched schemas. The Control Plane routes requests via a \textit{Routing Handler} using similarity search or rule-based matching. \\
\hline
\textbf{Structured Interfaces} & Both emphasize deterministic tool behavior through formalized I/O specifications to reduce prompt ambiguity. \\
\hline
\end{tabular}
\end{table}

\subsection{Key Differences Between the Control Plane and MCP}

While the Control Plane and MCP share structural themes, their operational design and goals diverge significantly. The Control Plane emphasizes orchestration, governance, and multi-agent extensibility, whereas MCP focuses on standardizing tool invocation for a single LLM context. Table \ref{tab:cp_mcp_differences}, shows how they differ from one another.

\begin{table}[ht]
\centering
\caption{Key Differences Between Control Plane and MCP}
\label{tab:cp_mcp_differences}
\begin{tabular}{|p{4.5cm}|p{5.8cm}|p{5.8cm}|}
\hline
\textbf{Aspect} & \textbf{Control Plane (This Work)} & \textbf{Model Context Protocol (MCP)} \\
\hline
\textbf{Architecture Type} & External modular orchestrator & Embedded schema-based interface \\
\hline
\textbf{Routing Strategy} & Rule-based and similarity-based routing via \textit{Routing Handler} & Implicit function selection via schema-matching in prompt \\
\hline
\textbf{Agent Scope} & Supports multiple agents and decoupled planning & Coupled to a single Claude-based LLM instance \\
\hline
\textbf{Governance and Tracking} & Includes \textit{Usage Tracker}, policy enforcement, failure handling & No built-in governance or logging \\
\hline
\textbf{Learning and Feedback} & Optional \textit{Feedback Integrator} for experience-based routing & No feedback loop or adaptive learning \\
\hline
\textbf{Tool Chaining Support} & Enables explicit chaining and dependency-based tool routing & No chaining logic; tools treated atomically \\
\hline
\textbf{Tool Fallback and Safety} & \textit{Failure Handler} supports default responses and recovery & No structured fallback mechanism \\
\hline
\textbf{Extensibility} & LLM-agnostic, framework-agnostic & Claude-specific runtime binding \\
\hline
\end{tabular}
\end{table}

\section{Conclusion and Future Directions}
The advent of generative models and their role in the development of Agentic AI systems, have also led to the rise of many frameworks. One of the challenges in developing systems the orchestration of the tools in an simple, safe, and manageable manner in production. The lack of composable, minimal design patterns is limiting the scalability of agentic AI. The proposed pattern was developed with that and model-agnosticism in mind. The proposed ``Control Plane as a Tool'' allows developers to encapsulate routing logic and enforce governance across environments. 

While this work addresses many of the current challenges in the development of Agentic AI systems, the pattern/approach has a potential to be extended to include many more features. Future work in this area will realize the development of a framework-agnostic system and evaluate performance, safety, and extensibility across larger multi-agent deployments.

\paragraph{Author Disclaimer.} This work was independently conceived and executed by the author without financial support from any institution, company, or donor organization. It was not funded by the author’s current employer or by any external grant, donation, or sponsorship. All opinions and technical claims are solely those of the author.

\paragraph{{Acknowledgements:}} This work would not have been possible without the unwavering support and understanding of my wife, Dharani, and my kids, Shree and Kart, even when I have to work late nights 



\bibliographystyle{plainnat}
\bibliography{references}

\end{document}